\documentclass[letterpaper]{sig-alternate-2013}

\usepackage{color}
\usepackage{multirow}
\usepackage{graphicx}
\usepackage{caption}
\usepackage{subcaption}
\usepackage{hyperref}

\permission{\copyright 2017 International World Wide Web Conference Committee \\ (IW3C2)
\copyrightetc{ACM \the\acmcopyr}
}

\begin{document}

\title{Dynamic Key-Value Memory Networks for\\ Knowledge Tracing}
%
%
%
%
%
\def\sharedaffiliation{%
\end{tabular}
\begin{tabular}{c}}
\newcommand\Mark[1]{\textsuperscript#1}

\numberofauthors{1}
\author{
	\alignauthor Jiani Zhang\Mark{{1,2}}, Xingjian Shi\Mark{3}, Irwin King\Mark{{1,2}}, Dit-Yan Yeung\Mark{3}\\ 
	\sharedaffiliation
	\affaddr{\Mark{1}Shenzhen Key Laboratory of Rich Media Big Data Analytics and Application,}  \\
	\affaddr{Shenzhen Research Institute, The Chinese University of Hong Kong, Shenzhen, China}   \\
	\affaddr{\Mark{2}Department of Computer Science and Engineering,}  \\
	\affaddr{The Chinese University of Hong Kong, Shatin, N.T., Hong Kong}\\
	\email{\{jnzhang, king\}@cse.cuhk.edu.hk} \\
	\\
	\affaddr{\Mark{3}Department of Computer Science and Engineering,}  \\
	\affaddr{Hong Kong University of Science and Technology, Kowloon, Hong Kong}\\
	\email{\{xshiab, dyyeung\}@cse.ust.hk}
}

\maketitle
\begin{abstract}
	{\em Knowledge Tracing} (KT) is a task of tracing evolving knowledge state of students with respect to one or more concepts as they engage in a sequence of learning activities. One important purpose of KT is to personalize the practice sequence to help students learn knowledge concepts efficiently.
	However, existing methods such as Bayesian Knowledge Tracing and Deep Knowledge Tracing either model knowledge state for each predefined concept separately or fail to pinpoint exactly which concepts a student is good at or unfamiliar with.
	To solve these problems, this work introduces a new model called Dynamic Key-Value Memory Networks (DKVMN) that can exploit the relationships between underlying concepts and directly output a student's mastery level of each concept.
	Unlike standard memory-augmented neural networks that facilitate a single memory matrix or two static memory matrices, our model has one static matrix called \emph{key}, which stores the knowledge concepts and the other dynamic matrix called \emph{value}, which stores and updates the mastery levels of corresponding concepts.
	Experiments show that our model consistently outperforms the state-of-the-art model in a range of KT datasets. Moreover, the DKVMN model can automatically discover underlying concepts of exercises typically performed by human annotations and depict the changing knowledge state of a student.
\end{abstract}


\section{Introduction}
With the advent of massive open online courses and intelligent tutoring systems in the web, students can get appropriate guidance and acquire relevant knowledge in the process of solving exercises.
When an exercise is posted, a student must apply one or more concepts to solve the exercise. For example, when a student attempts to solve the exercise ``1+2'', then he or she should apply the concept of ``integer addition''; when a student attempts to solve ``1+2+3.4'', then he or she should apply the concepts of ``integer addition'' and ``decimal addition''. The probability that a student can answer the exercise correctly is based on the student's {\em knowledge state}, which stands for the depth and robustness of the underlying concepts the student has mastered. 

The goal of {\em knowledge tracing} (KT) is to trace the knowledge state of students based on their past exercise performance.  
KT is an essential task in online learning platforms.
Tutors can give proper hints and tailor the sequence of practice exercises based on the personal strengths and weaknesses of students.
Students can be made aware of their learning progress and may devote more energy to less-familiar concepts to learn more efficiently.

Although effectively modeling the knowledge of students has high educational impact, using numerical simulations to represent the human learning process is inherently difficult~\cite{piech2015deep}.
Usually, KT is formulated as a supervised sequence learning problem:
given a student's past exercise interactions $\mathcal{X}= \{\mathbf{x}_1$,$\mathbf{x}_2$,...,$\mathbf{x}_{t-1}\}$, predict the probability that the student will answer a new exercise correctly, i.e., $p(r_t=1|q_t,\mathcal{X})$.
Input $\mathbf{x}_{t}=(q_{t},r_{t})$ is a tuple containing the exercise $q_{t}$, which student attempts at the timestamp $t$, and the correctness of the student's answer $r_{t}$.
We model $\mathcal{X}$ as observed variables and a student's knowledge state $\mathcal{S}= \{\mathbf{s}_1$,$\mathbf{s}_2$,...,$\mathbf{s}_{t-1}\}$ of $N$ underlying concepts $\mathcal{C}= \{c^1,c^2,...,c^N\}$ as a hidden process.

Existing methods such as Bayesian Knowledge Tracing (BKT)~\cite{corbett1994knowledge} and Deep Knowledge Tracing (DKT)~\cite{piech2015deep} model the knowledge state of students either in a concept specific manner or in one summarized hidden vector, as shown in Figure~\ref{fig:introduction}.
In BKT, a student's {\em knowledge state} $\mathbf{s}_t$ is analyzed into different {\em concept state}s $\{\mathbf{s}_t^i\}$ and BKT models each concept state separately. 
BKT assumes the concept state as a binary latent variable, {\em known} and {\em unknown}, and uses a Hidden Markov model to update the posterior distribution of the binary concept state. 
Therefore,  BKT cannot capture the relationship between different concepts. 
Moreover, to keep the Bayesian inference tractable, BKT uses discrete random variables and simple transition models to describe the evolvement of each concept state.
As a result, although BKT can output the student's mastery level of some predefined concepts, it lacks the ability to extract undefined concepts and model complex concept state transitions.

\begin{figure}[!tb]
	\centering
	\includegraphics[width=0.47\textwidth]{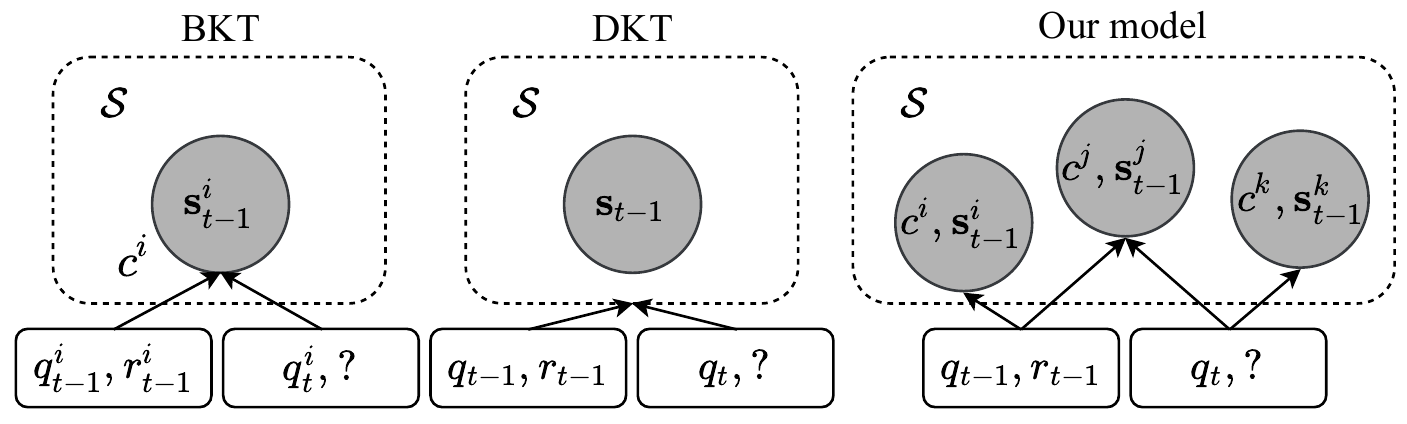}
	\caption{Model differences among BKT, DKT, and our model. BKT is concept specific. DKT uses a summarized hidden vector to model the knowledge state. Our model maintains the {\em concept state} for each concept simultaneously and all concept states constitute the {\em knowledge state} of a student.}
	\label{fig:introduction}
\end{figure}

Besides solving the problem from the Bayesian perspective, a deep learning method named DKT~\cite{piech2015deep} exploits a variant of recurrent neural networks (RNNs) called long short-term memory (LSTM)~\cite{hochreiter1997long}. 
LSTM assumes a high-dimensional and continuous representation of the underlying knowledge state $\mathcal{S}$.
The nonlinear input-to-state and state-to-state transitions of DKT have stronger representational power than those of BKT. No human-labeled annotation is required. However, DKT summarizes a student's knowledge state of all concepts in one hidden state, which makes it difficult to trace how much a student has mastered a certain concept and pinpoint which concepts a student is good at or unfamiliar with~\cite{khajahdeep,wilsonestimating}.

The present work introduces a new model called Dynamic Key-Value Memory Networks (DKVMN) that combines the best of two worlds: the ability to exploit the relationship between concepts and the ability to trace each concept state. 
Our DKVMN model can automatically learn the correlation between input exercises and underlying concepts and maintain a concept state for each concept. At each timestamp, only related concept states will be updated. 
For instance, in Figure~\ref{fig:introduction}, when a new exercise $q_t$ comes, the model finds that $q_t$ requires the application of concept $c^j$ and $c^k$. Then we read the corresponding concept states $\mathbf{s}^j_{t-1}$ and $\mathbf{s}^k_{t-1}$ to predict whether the student will answer the exercise correctly. After the student completes the exercise, our model will update these two concept states. All concept states constitute the {\em knowledge state} $\mathcal{S}$ of a student. 

In addition, unlike standard memory-augmented neural networks (MANNs) that facilitate a single memory matrix~\cite{graves2014neural,santoro2016meta,weston2014memory} or a variation with two static memory matrices~\cite{miller2016key,sukhbaatar2015end}, our model has one static matrix called \emph{key}, which stores the concept representations and the other dynamic matrix called \emph{value}, which stores and updates the student's understanding (concept state) of each concept. The terms {\em static} and {\em dynamic} matrices are respectively analogous to {\em immutable} and {\em mutable} objects as keys and values in the dictionary data structure (e.g., Python's dictionary). Meanwhile, our training process is analogous to object creation. After the keys are created, they will be fixed (i.e., immutable) during testing.

The network with two static memory matrices is not suitable for solving the KT task because learning is not a static process. Learning builds upon and is shaped by previous knowledge in human memory~\cite{grefenstette2015learning}. The model with a single dynamic matrix maps the exercise with the correct answer and the exercise with the incorrect answer to different concept states, which does not match our cognition. Experiments show that our DKVMN model outperforms the MANN model with a single memory matrix and the state-of-the-art model.

Our main contributions are summarized as follows:
\begin{enumerate}
	\item The utility of MANNs is exploited to better simulate the learning process of students.
	\item A novel DKVMN model with one static \emph{key} matrix and one dynamic \emph{value} matrix is proposed.
	\item Our model can automatically discover concepts, a task that is typically performed by human experts, and depict the evolving knowledge state of students.
	\item Our end-to-end trainable model consistently outperforms BKT and DKT on one synthetic and three real-world datasets respectively.
\end{enumerate}

\section{Related Works}
\subsection{Knowledge Tracing}

The KT task evaluates the knowledge state of a student based simply on the correctness or incorrectness $r_t$ of a student's answers in the process of solving exercises $q_t$. In this study $q_t$ is an exercise tag and $r_t \in \{0,1\}$ is a binary response (1 is correct and 0 is incorrect).
No secondary data are incorporated~\cite{khajahdeep}.

BKT~\cite{corbett1994knowledge} is a highly constrained and structured model~\cite{khajahdeep} because it models concept-specific performance, i.e., an individual instantiation of BKT is made for each concept, and BKT assumes knowledge state as a binary variable.
Many following variations were raised by integrating personalization study~\cite{pardos2010modeling,yudelson2013individualized}, exercise diversity~\cite{pardos2011kt}, and other information~\cite{d2008contextual,reye2004student} into the Bayesian framework.

DKT~\cite{piech2015deep} exploits the utility of LSTM~\cite{hochreiter1997long} to break the restriction of skill separation and binary state assumption. 
LSTM uses hidden states as a kind of summary of the past sequence of inputs, and the same parameters are shared over different time steps.
Experiments in~\cite{piech2015deep} showed that DKT outperforms previous Bayesian models by a large margin in terms of prediction accuracy.
This study was the first attempt to integrate deep learning models~\cite{lecun2015deep,schmidhuber2015deep}, which have achieved significant success in other areas, including computer vision~\cite{krizhevsky2012imagenet} and natural language processing~\cite{mikolov2013distributed} into KT.

\subsection{Memory-Augmented Neural Networks}

Inspired by computer architecture, a particular neural network module called external memory was proposed to enhance the ability of a network to capture long-term dependencies and solve algorithmic problems~\cite{graves2016hybrid}. MANN have led the progress in various areas, such as question answering~\cite{weston2014memory,sukhbaatar2015end,bordes2015large,miller2016key}, natural language transduction~\cite{grefenstette2015learning}, algorithm inference~\cite{graves2014neural,joulin2015inferring}, and one-shot learning~\cite{santoro2016meta,vinyals2016matching}.

The typical external memory module contains two parts, a memory matrix that stores the information and a controller that communicates with the environment and reads or writes to the memory. The reading and writing operations are achieved through additional attention mechanisms. Most previous works~\cite{graves2014neural,sukhbaatar2015end,santoro2016meta} use a similar way to compute the read weight. For an input $\mathbf{k}_t$, a cosine similarity or an inner product $K[\mathbf{k}_t, \mathbf{M}_t(i)]$ of the input and each memory slot $\mathbf{M}_t(i)$ is computed,
which then goes through a softmax with a positive key strength $\beta_t$ to obtain a read weight $\mathbf{w}^r_t$ :
$w_t^r(i)= \text{Softmax}(\beta_t K[\mathbf{k}_t, \mathbf{M}_t(i)]) $, where Softmax($z_i$) = $e^{z_i}/\sum_j e^{z_j}$. For the write process, an attention mechanism of focusing both by content and by location is proposed in~\cite{graves2014neural} to facilitate all the locations of the memory. In addition, a pure content-based memory writer named least recently used access (LRUA) module is raised in~\cite{santoro2016meta} to write the key either to the least recently used memory location or to the most recently used memory location.

\begin{figure*}[!tb]
	\centering
	\begin{subfigure}[b]{0.48\textwidth}
		\centering
		\includegraphics[width=0.9\textwidth]{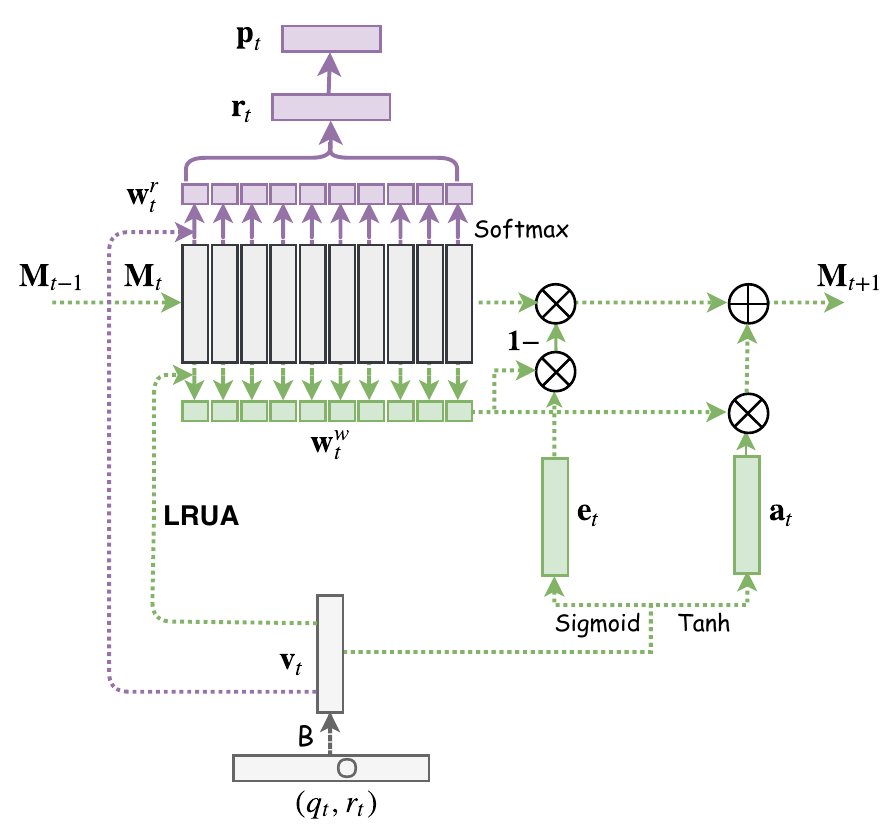}
		\caption{Architecture for Memory-Augmented Neural Networks.}
		\label{fig:MANN}
	\end{subfigure}
	~ 
	\begin{subfigure}[b]{0.48\textwidth}
		\centering
		\includegraphics[width=0.85\textwidth]{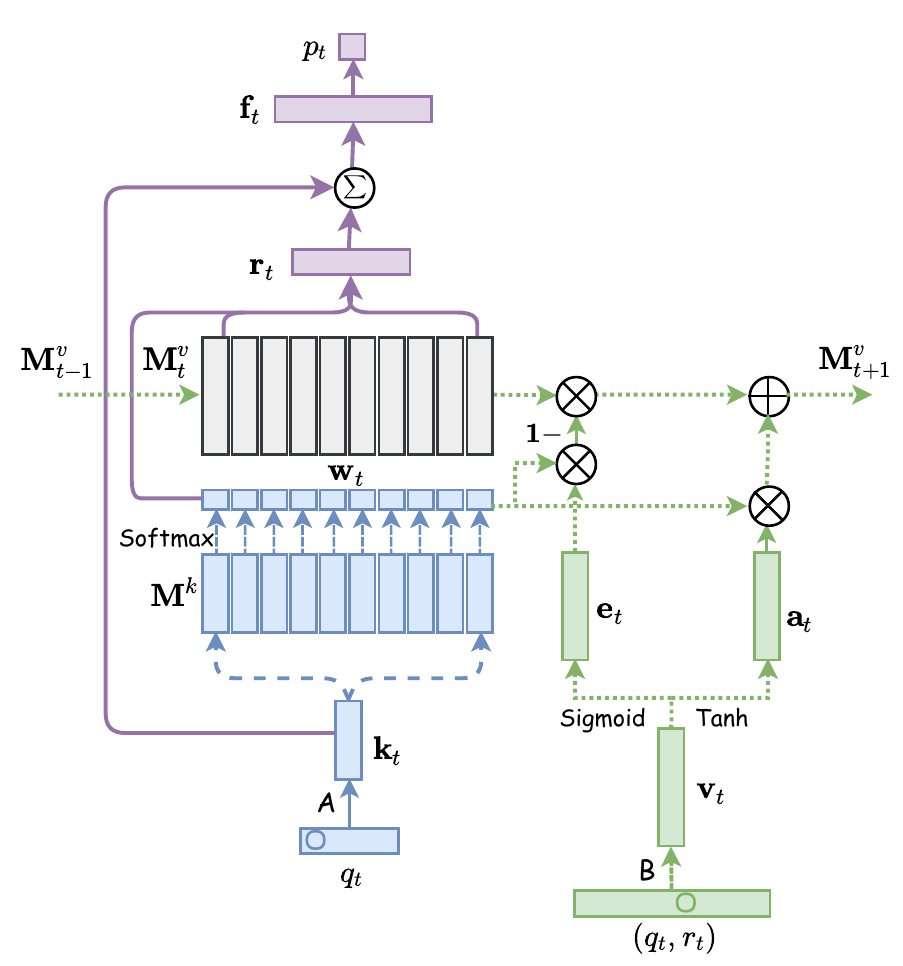}
		\caption{Architecture for Dynamic Key-Value Memory Networks.}
		\label{fig:KVMN}
	\end{subfigure}
	\caption{In both architecture, the model is only drawn at the timestamp t, where the purple components describe the read process and the green components describe the write process. The blue components in the DKVMN model denote the attention process to compute the corresponding weight. (Best viewed in color.)}
	\label{fig:Architecture}
\end{figure*}

Owing to the recurrence introduced in the read and write operations, MANN is a special kind of RNN as well. However, MANN is different from conventional RNNs like the LSTM used in DKT in three aspects. First, traditional RNN models use a single hidden state vector to encode the temporal information, whereas, MANN uses an external memory matrix that can increase storage capacity~\cite{weston2014memory}. Second, the state-to-state transition of traditional RNNs is unstructured and global, whereas MANN uses read and write operations to encourage local state transitions~\cite{graves2014neural}. Third, the number of parameters in traditional RNNs is tied to the size of hidden states~\cite{santoro2016meta}. For MANN, increasing the number of memory slots will not increase the number of parameters, an outcome that is more computationally efficient.

\section{Model}
In this section, we first introduce the way to exploit the existing MANN model to solve the KT problem. We then show the deficiencies of MANN and describe our DKVMN model.
In our description below, we denote vectors with bold small letters and matrices with bold capital letters.

\subsection{Memory-Augmented Neural Network for Knowledge Tracing}

To solve the KT problem, the external memory matrix of MANN is treated as the knowledge state of a student. The overall structure of the model is shown in Figure~\ref{fig:MANN}. The memory, denoted as $\mathbf{M}_t$, is an $N \times M$ matrix, where $N$ is the number of memory locations, and $M$ is the vector size at each location.
At each timestamp $t$, the input for MANN is a joint embedding $\mathbf{v}_t$ of $(q_t,r_t)$, where each $q_t$ comes from a set of $Q$ distinct exercise tags and $r_t$ is a binary value indicating whether the student answered the exercise correctly.
The embedding vector $\textbf{v}_t$ is used to compute the read weight $\textbf{w}^r_t$ and the write weight $\textbf{w}^w_t$. 

In our implementation, we choose the cosine similarity attention mechanism to compute $\textbf{w}^r_t$ and the LRUA mechanism~\cite{santoro2016meta} to compute $\textbf{w}^w_t$. Details of these two attention mechanisms are shown in the appendix.
The intuition of the MANN is that when a student answers the exercise that has been stored in the memory with the same response, $\textbf{v}_t$ will be written to the previously used memory locations and when a new exercise arrives or the student gets a different response, $\textbf{v}_t$ will be written to the least recently used memory locations.

In the read process, the read content $\mathbf{r}_t$ is obtained by the weighted sum of all memory slots with the read weight $\mathbf{w}^r_t$:
\begin{equation}
\mathbf{r}_t = \sum_{i=1}^N w_t^r(i) \mathbf{M}_t(i).
\end{equation}
The output $\textbf{p}_t \in \mathbb{R}^Q$, which is computed from $\textbf{r}_t$, indicates the probability that the student can answer each exercise correctly in the next timestamp. 

In the write process, we first erase unnecessary contents in the memory using the erase signal $\textbf{e}_t$ and the write weight $\textbf{w}^r_t$, and then add $\textbf{v}_t$ into the memory using the add signal $\textbf{a}_t$~\cite{graves2014neural}. For further details, see Section 3.2.3.

MANN uses $N$ memory slots to encode the knowledge state of a student and has a larger capacity than LSTM, which only encodes knowledge state in a single hidden vector.

\subsection{Dynamic Key-Value Memory Networks}
Despite being more powerful than LSTM in storing the past performance of students, MANN still has deficiencies when applied to the KT task. In MANN, the content we read lies in the same space as the content we write. However, for tasks like KT, the input and the prediction, which are the exercises the student receives and the correctness of the student's answer have different types. Therefore, the way to embed the exercise and the response jointly as the attention key does not make sense. Furthermore, MANN cannot explicitly model the underlying concepts for input exercises. Knowledge state of a particular concept is dispersed and cannot be traced.

To solve these problems, our DKVMN model uses key-value pairs rather than a single matrix for the memory structure.
Instead of attending, reading, and writing to the same memory matrix in MANN, our DKVMN model attends input to the {\em key} component, which is immutable, and reads and writes to the corresponding {\em value} component.

Unlike MANN, at each timestamp, DKVMN takes a discrete exercise tag $q_t$, outputs the probability of response $p(r_t|q_t)$, and then updates the memory with exercise and response tuple $(q_t,r_t)$.
Here, $q_t$ also comes from a set with $Q$ distinct exercise tags and $r_t$ is a binary value. We further assume there are $N$ latent concepts $\{c^1, c^2,...,c^N\}$ underlying the exercises. These concepts are stored in the {\em key} matrix $\mathbf{M}^k$ (of size $N \times d_k$) and the student's mastery levels of each concept, i.e., concept states $ \{\mathbf{s}^1_{t},\mathbf{s}^2_{t},...,\mathbf{s}^N_{t}\}$ are stored in the {\em value} matrix $\mathbf{M}_t^v$ (of size $N \times d_v$), which changes over time.

DKVMN traces the knowledge of a student by reading and writing to the \emph{value} matrix using the correlation weight computed from the input exercise and the \emph{key} matrix. The model details are elaborated in the following sections.

\subsubsection{Correlation Weight}

The input exercise $q_t$ is first multiplied by an embedding matrix $\mathbf{A}$ (of size $Q \times d_k$) to get a continuous embedding vector $\mathbf{k}_t$ of dimension $d_k$.
The correlation weight is further computed by taking the softmax activation of the inner product between $\mathbf{k}_t$ and each {\em key} slot $\mathbf{M}^k(i)$:
\begin{equation}
w_t(i) = \text{Softmax}(\mathbf{k}_t^T \mathbf{M}^k(i)),
\end{equation}
where Softmax($z_i$) = $e^{z_i}/\sum_j e^{z_j}$ and is differentiable.
Both the read and write processes will use this weight vector $\mathbf{w}_t$, which represents the correlation between exercise and each latent concept.

\subsubsection{Read process}

When an exercise $q_t$ comes, the read content $\mathbf{r}_t$ is retrieved by the weighted sum of all memory slots in the {\em value} matrix using $\mathbf{w}_t$:
\begin{equation} \label{eq:r_t}
\mathbf{r}_t = \sum_{i=1}^N w_t(i) \mathbf{M}^v_t(i).
\end{equation}
The calculated read content $\mathbf{r}_t$ is treated as a summary of the student's mastery level of this exercise.
Given that each exercise has its own difficulty, we concatenate the read content $\mathbf{r}_t$ and the input exercise embedding $\mathbf{k}_t$ and then pass it through a fully connected layer with a Tanh activation to get a summary vector $\mathbf{f}_t$, which contains both the student's mastery level and the prior difficulty of the exercise:
\begin{equation} \label{eq:f_t}
\mathbf{f}_t = \text{Tanh}(\mathbf{W}_1^T [\mathbf{r}_t,\mathbf{k}_t]+\mathbf{b}_1),
\end{equation}
where Tanh($z_i$) = $(e^{z_i}-e^{-z_i})/(e^{z_i}+e^{-z_i})$.

Finally, $\mathbf{f}_t$ is passed through another fully connected layer with a Sigmoid activation to predict the performance of the student:
\begin{equation} \label{eq:p_t}
p_t = \text{Sigmoid}(\mathbf{W}_2^T \mathbf{f}_t+\mathbf{b}_2),
\end{equation}
where Sigmoid($z_i$) = $1/(1+ e^{-z_i})$, and $p_t$ is a scalar that represents the probability of answering $q_t$ correctly.

\begin{table*}[!tb]
	\centering
	\caption{Test AUC results for all datasets. BKT is the standard BKT. BKT+ is the best-reported result with BKT variations. DKT is the result using LSTM. MANN is the baseline using a single memory matrix. DKVMN is our model.}
	\begin{tabular}{c  rrrr r rrrr}
		\hline
		\multirow{2}{*}{Datasets} & \multicolumn{3}{c}{Overview} & & \multicolumn{5}{c}{Test AUC (\%)} \\
		\cline{2-4}  \cline{6-10} 
		& Students & Exercise Tags & Records & & BKT & BKT+ & DKT & MANN & DKVMN \\
		\hline \hline
		Synthetic-5    & 4,000 & 50   & 200,000 & & 62 & 80 & 80.3$\pm$0.1 & 81.0$\pm$0.1 & \textbf{82.7$\pm$0.1} \\ \hline
		ASSISTments2009& 4,151 & 110  & 325,637 & & 63 & -  & 80.5$\pm$0.2 & 79.7$\pm$0.1   & \textbf{81.6$\pm$0.1}\\ \hline
		ASSISTments2015& 19,840& 100  & 683,801 & & 64 & -  & 72.5$\pm$0.1 & 72.3$\pm$0.2 & \textbf{72.7$\pm$0.1} \\ \hline
		Statics2011    & 333   & 1,223& 189,297 & & 73 & 75 & 80.2$\pm$0.2 & 77.6$\pm$0.1 & \textbf{82.8$\pm$0.1} \\ \hline
	\end{tabular}
	\label{table:result}
\end{table*}

\subsubsection{Write process}

After the student answers the question $q_t$, the model will update the {\em value} matrix according to the correctness of the student's answer. A joint embedding of $(q_t,r_t)$ will be written to the {\em value} part of the memory with the same correlation weight $\mathbf{w}_t$ used in the read process.

The tuple $(q_t,r_t)$ is embedded with an embedding matrix $\mathbf{B}$ of size $2Q \times d_v$ to obtain the {\em knowledge growth} $\mathbf{v}_t$ of the student after working on this exercise.
When writing the student's {\em knowledge growth} into the {\em value} component, the memory is erased first before new information is added~\cite{graves2014neural}, a step inspired by the input and forget gates in LSTMs.

Given a write weight (which is the correlation weight $\mathbf{w}_t$ in our model), an {\em erase vector} $\mathbf{e}_t$ is computed from $\mathbf{v}_t$:
\begin{equation}
\mathbf{e}_t = \text{Sigmoid}(\mathbf{E}^T\mathbf{v}_t + \mathbf{b}_e),
\end{equation}
where the transformation matrix $\mathbf{E}$ is of shape $d_v \times d_v$,  $\mathbf{e}_t$  is a column vector with $d_v$ elements that all lie in the range $(0,1)$. The memory vectors of {\em value} component $\mathbf{M}^v_{t-1}(i)$ from the previous timestamp are modified as follows:
\begin{equation}
\tilde{\mathbf{M}}^v_t(i) = \mathbf{M}^v_{t-1}(i)[\mathbf{1} - w_t(i) \mathbf{e}_t] ,
\end{equation}
where $\mathbf{1}$ is a row-vector of all 1-s. Therefore, the elements of a memory location are reset to zero only if both the weight at the location and the erase element are one. The memory vector is left unchanged if either the weight or the erase signal is zero.

After erasing, a length $d_v$ {\em add vector} $\mathbf{a}_t$ is used to update each memory slot: 
\begin{equation}
\mathbf{a}_t = \text{Tanh}(\mathbf{D}^T \mathbf{v}_t + \mathbf{b_a})^T,
\end{equation}
where the transformation matrix $\mathbf{D}$ is of shape $d_v \times d_v$ and $\mathbf{a}_t$ is a row vector. The {\em value} memory is updated at each time $t$ by
\begin{equation}
\mathbf{M}^v_t(i) = \tilde{\mathbf{M}}^v_{t-1}(i) + w_t(i)\mathbf{a}_t.
\end{equation}

This {\em erase}-followed-by-{\em add} mechanism allows forgetting and strengthening concept states in the learning process of a student.

%

\subsubsection{Training}

The overall model architecture is shown in Figure~\ref{fig:KVMN}. 
During training, both embedding matrices $\mathbf{A}$ and $\mathbf{B}$, as well as other parameters and the initial value of $\mathbf{M}^k$ and $\mathbf{M}^v$ are jointly learned by minimizing a standard cross entropy loss between $p_t$ and the true label $r_t$.
\begin{equation}
\mathcal{L} = -\sum_t (r_t\text{log}p_t + (1-r_t)\text{log}(1-p_t)).
\end{equation}

Our DKVMN model is fully differentiable and can be trained efficiently with stochastic gradient descent (see Section 4.2 for more details).


\section{Experiments}

The prediction accuracy is first evaluated by comparing our DKVMN model with other methods on four datasets, namely one synthetic dataset and three real-world datasets, collected from online learning platforms.
Then, comparative experiments of different dimensions of states are performed on DKVMN and DKT for further model exploration.
Finally, the ability of our model is verified to discover concepts automatically and depict the knowledge state of students.

The experiment results lead to the following findings:
\begin{itemize}
	\item DKVMN outperforms the standard MANN and the state-of-the-art method on four datasets.
	\item DKVMN can produce better results with fewer parameters than DKT.
	\item DKVMN does not suffer from overfitting, which is a big issue for DKT.
	\item DKVMN can discover underlying concepts for input exercises precisely.
	\item DKVMN can depict students' concept states of distinct concepts over time.
\end{itemize}
We implement the models using MXNet~\cite{chen2015mxnet} on a computer with a single NVIDIA K40 GPU.

\subsection{Datasets}

To evaluate performance, we test KT models on four datasets: Synthetic-5, ASSISTments2009, ASSISTments2015, and Statics2011. 

\textbf{Synthetic-5}: This dataset\footnote{Synthetic-5:\url{https://github.com/chrispiech/DeepKnowledgeTracing/tree/master/data/synthetic}} simulates 2000 virtual students answering 50 exercises in both the training and testing dataset. Each exercise is drawn from one of five hidden concepts and has different levels of difficulty. We have no access to the underlying concept labels in the training process and simply use them as the ground truth to evaluate the discovered concept results using our DKVMN model.

\textbf{ASSISTments2009}: This dataset~\cite{feng2009addressing} is gathered from the ASSISTments online tutoring platform.
Owing to duplicated record issues~\cite{xionggoing}, an updated version is released and all previous results on the old dataset are no longer reliable. The experiments in our paper are conducted using the updated ``skill-builder'' dataset\footnote{ASSISTments2009:\url{https://sites.google.com/site/assistmentsdata/home/assistment-2009-2010-data/skill-builder-data-2009-2010}}. Records without skill names are discarded in the preprocessing. Thus, the number of records in our experiments is smaller than that in~\cite{xionggoing}. A total of
4,151 students answer 325,637 exercises along with 110 distinct exercise tags.

\textbf{ASSISTments2015}: ASSISTments2015\footnote{ASSISTments2015:\url{https://sites.google.com/site/assistmentsdata/home/2015-assistments-skill-builder-data}} only contains student responses on 100 skills.
After preprocessing (removing the value of {\em correct} $\notin \{0,1\}$), 683,801 effective records from 19,840 students are remained in this dataset. Each problem set in this dataset has one associated skill. Although this dataset has the largest number of records, the average records for each student are the lowest.

\textbf{Statics2011}: Statics\footnote{Statics2011:\url{https://pslcdatashop.web.cmu.edu/DatasetInfo?datasetId=507}} is from a college-level engineering statics course with 189,297 trials, 333 students and 1,223 exercises tags~\cite{Steif2011STATICS,koedinger2010data}. In our experiments, a concatenation of problem name and step name is used as an exercise tag; thus it has the maximum number of exercise tags and the maximum number of average records per student.

The complete statistical information for all datasets can be found in Table~\ref{table:result}.

\begin{table*}[tb!]
	\centering
	\caption{Comparison of DKVMN with DKT on four datasets with different numbers of state dimensions and memory size $N$. \\``s.~dim'', ``m.~size'' and ``p.~num'' represent the state dimension, memory size (i.e., the number of concepts $N$), and the number of parameters, respectively. We choose the state dimensions of 10, 50, 100, and 200 for both DKT and DKVMN. Then for DKVMN, we change memory size for 1, 2, 5, 10, 20, 50, and 100 for each state dimension and report the best test AUC with the corresponding memory size. We likewise compare the number of parameters for both models.}
	\begin{tabular}{|c | rrrr | rrrr | rrrr | rrrr|}
		\hline
		\multirow{3}{*}{Model} & \multicolumn{4}{c|}{Synthetic-5} & \multicolumn{4}{c|}{ASSISTments2009}& \multicolumn{4}{c|}{ASSISTments2015} & \multicolumn{4}{c|}{Statics2011} \\ 
		& s. & m. & test & p. & s. & m. & test & p. & s. & m. & test & p. & s. & m. & test & p.  \\ 
		& dim & size & auc & num & dim & size & auc & num & dim & size & auc & num & dim & size & auc & num\\
		\hline \hline
		\multirow{4}{*}{DKT}  
		& 10  & -  & 80.06 & 2.4K      & 10  & - & 80.38 & 4.3K      & 10  & - & 72.40 & 4.0K & 10  & - &  78.12 & 39K \\
		& 50  & -  & 80.22 & 28K       & 50  & - & \textbf{80.53}&37K& 50  & - & \textbf{72.52} & 36K & 50  & - & 79.86 & 205K \\
		& 100 & -  & \textbf{80.34}&96K& 100 & - & 80.51 & 114K      & 100 & - &  72.49 & 111K & 100 & - & 80.16& 449K \\
		& 200 & -  & 80.32 & 352K      & 200 & - & 80.43 & 388K      & 200 & - & 72.45 & 382K & 200 & - & \textbf{80.20} & 1.0M \\
		\hline \hline
		\multirow{4}{*}{DKVMN} 
		& 10 & 50 & 82.00 & 12K       & 10 & 10 & 81.47 & 7k & 10 & 20 & \textbf{72.68} & 14K & 10 & 10 &82.72 & 92K\\ 
		& 50 & 50 & 82.66 & 25K       & 50 & 20 & \textbf{81.57} & 31k & 50 & 10 & 72.66 & 29K & 50 & 10 & \textbf{82.84} & 197K\\
		& 100& 50 & \textbf{82.73}&50K& 100& 10 & 81.42 & 68k & 100& 50 & 72.64 & 63K & 100 & 10 & 82.71 & 338K\\
		& 200& 50 & 82.71 & 130K      & 200& 20 & 81.37 & 177k & 200& 50 & 72.53 & 153K & 200& 10 & 82.70 & 649K\\
		\hline
	\end{tabular}
	\label{table:result_KVMN_DKT}
\end{table*}

\begin{figure*}[tb!]
	\centering
	\includegraphics[width=0.24\textwidth]{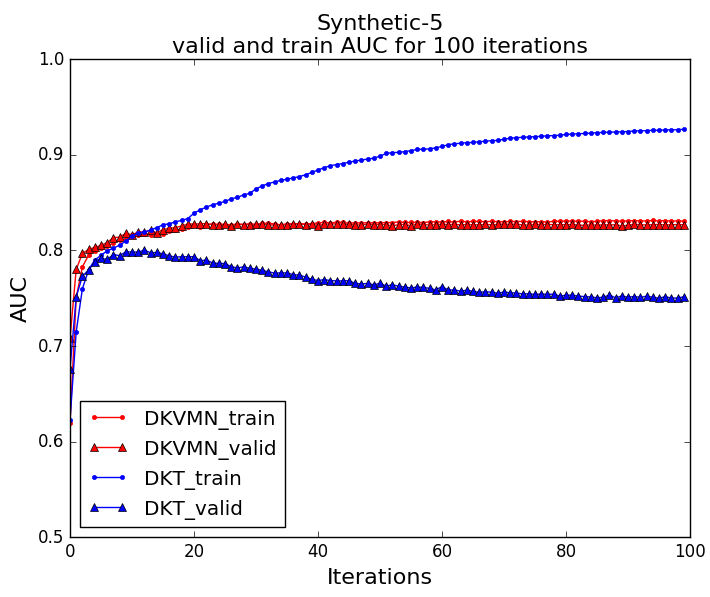}
	\includegraphics[width=0.24\textwidth]{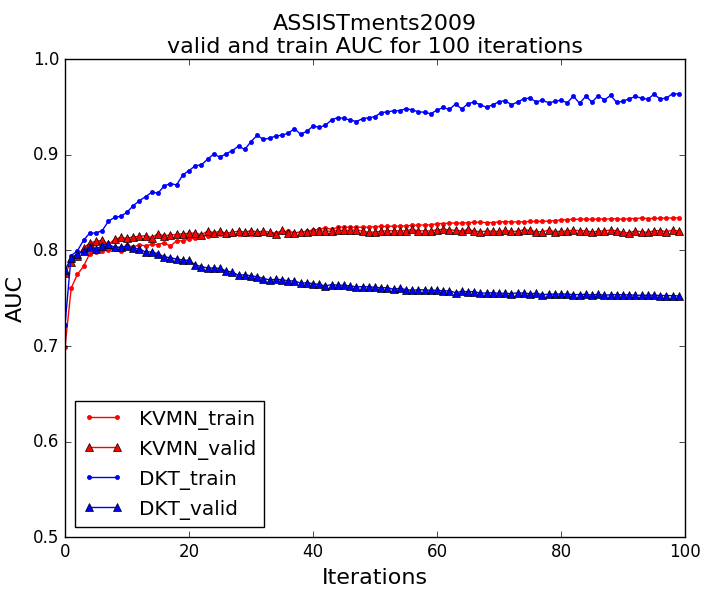}
	\includegraphics[width=0.24\textwidth]{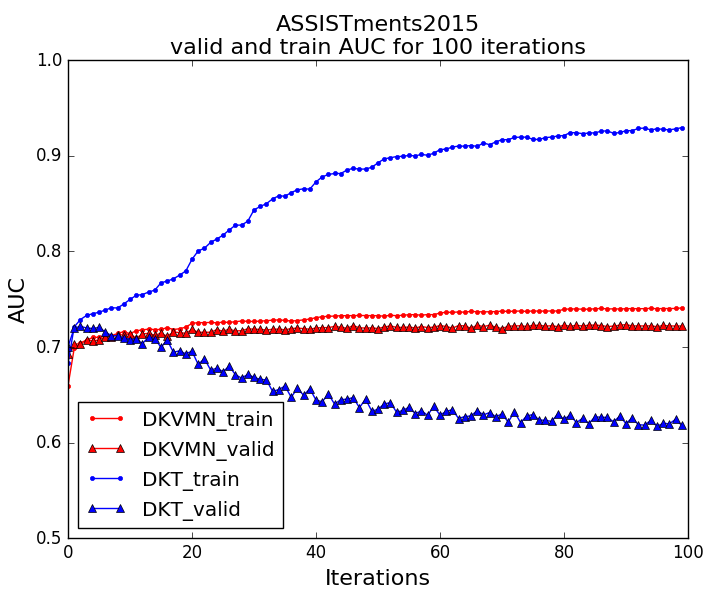}
	\includegraphics[width=0.24\textwidth]{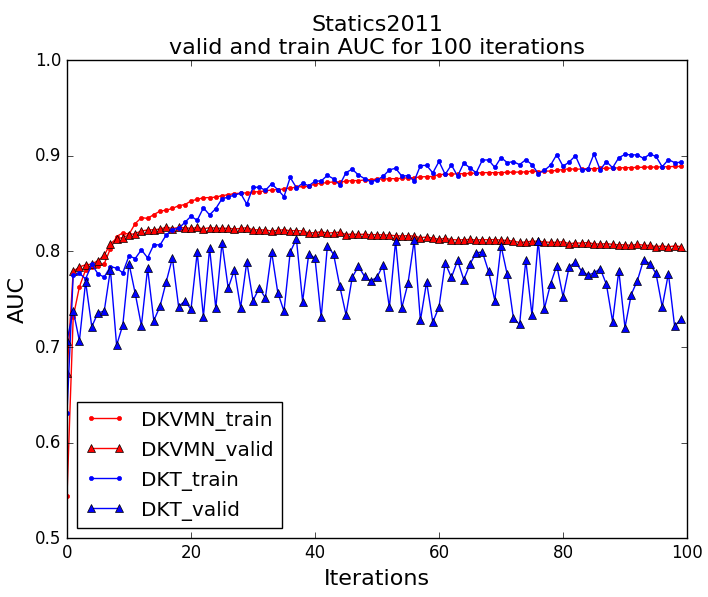}
	\caption{Validation AUC and training AUC of DKVMN and DKT on all datasets. The blue line represents the DKT model, and the red line represents our DKVMN model. The dotted line represents the training AUC and the line with upper triangles represents the validation AUC. (Best viewed in color.)}
	\label{fig:AUC}
\end{figure*}

\subsection{Implementation Details}
The input exercise data are presented to neural networks using ``one-hot'' input vectors. Specifically, if $Q$ different exercises exist in total, then the exercise tag $q_t$ for the {\em key} memory part is a length $Q$ vector whose entries are all zero except for the $q_t^{th}$ entry, which is one. Similarly, the combined input $x_t =(q_t,r_t)$ for the {\em value} matrix component is a length $2Q$ vector, where entry $x_t = q_t + r_t * Q$ is one.

We learn the initial value of both the {\em key} and the {\em value} matrix in the training process.
Each slot of the {\em key} memory is the concept embedding and is fixed in the testing process. Meanwhile, the initial value of the {\em value} memory is the initial state of each concept, which represents the initial difficulty of each concept.

Of all the datasets, 30\% of the sequences were held out as a testing set, except for the synthetic dataset where training and testing datasets had the same size. A total 20\% of the training set was split to form a validation set, which was used to select the optimal model architecture and hyperparameters and perform early stopping~\cite{orr2003neural}.

The parameters were initialized randomly from a Gaussian distribution with zero mean and standard deviation $\sigma$. 
The initial learning rate was case by case because the number of students, exercise tags, and total answers per dataset varied, but the learning rate $\gamma$ annealed every 20 epochs by $\gamma /1.5$ until the 100th epoch was reached. 

We used LSTM for DKT in our implementation. The standard MANN was implemented using the cosine similarity reading attention mechanism and the LRUA writing attention mechanism. Stochastic gradient descent with momentum and norm clipping~\cite{pascanu2013difficulty} were used to train DKT, MANN, and our DKVMN in all the experiments. We consistently set the momentum to be $0.9$ and the norm clipping threshold to be $50.0$. 
Given that input sequences are of different lengths, all sequences were set to be a length of $200$ (for {\em synthetic} with a length of $50$) and a null symbol was used to pad short sequence to a fixed size of $200$. 

In all cases, hyperparameters were tuned using the five-fold cross validation. The test area under the curve (AUC) was computed using the model with the highest validation AUC among the 100 epochs.
We repeated each training five times with different initializations $\sigma$ and reported the average test AUC along with the standard deviation.

\subsection{Student Performance Prediction}

\begin{figure*}[!tb]
	\centering
	\includegraphics[width=0.54\textwidth]{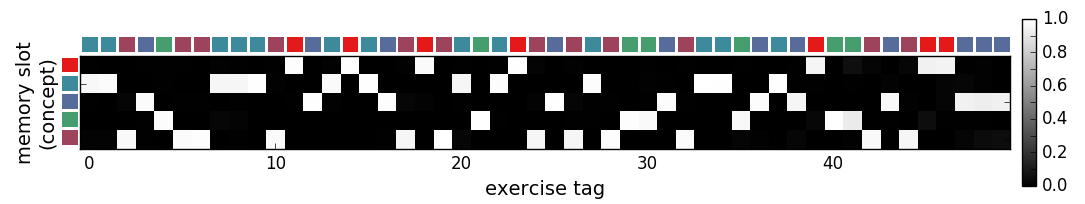}
	\includegraphics[width=0.08\textwidth]{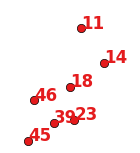}
	\includegraphics[width=0.08\textwidth]{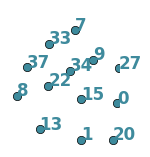}
	\includegraphics[width=0.09\textwidth]{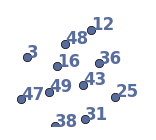}
	\includegraphics[width=0.09\textwidth]{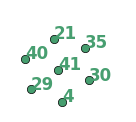}
	\includegraphics[width=0.09\textwidth]{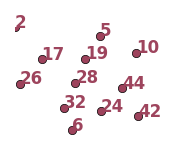}
	\caption{Concept discovery results on the synthetic-5 dataset when the memory size $N$ is set to be 5. In the left heat map, the x-axis represents each exercise and the y-axis represents the correlation weight between the exercise and five latent concepts generated from our DKVMN model. The ground-true concept is labeled on the top of each exercise. In the right exercise clustering graph, each node number represents an exercise. Exercises from the same ground-truth concept are clustered together. (Best viewed in color.)}
	\label{fig:exe_cluster_synthetic}
\end{figure*}

\begin{figure*}[!tb]
	\centering
	\includegraphics[width=0.29\textwidth]{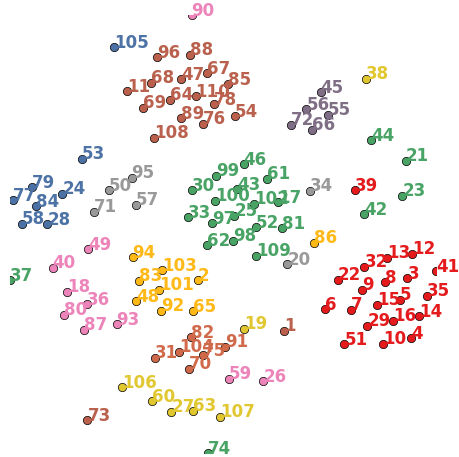}
	\quad
	\includegraphics[width=0.68\textwidth]{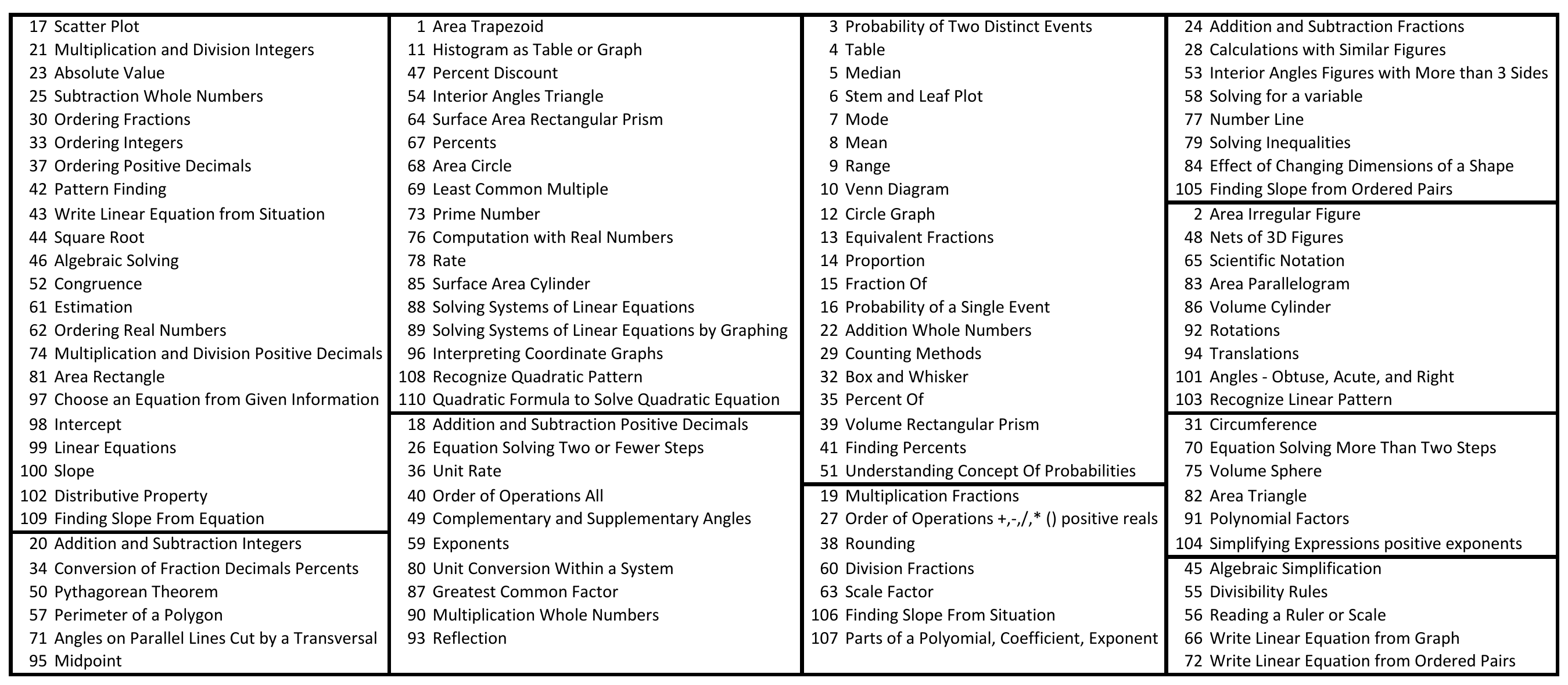}
	\caption{Concept discovery results on the ASSSISTments2009 dataset. 110 exercises are clustered into ten concepts. Exercises under the same concept are labeled in the same color in the left picture and also are put in the same block in the right table. (Best viewed in color.)}
	\label{fig:exe_cluster_assistments2009}
\end{figure*}

The AUC is measured to evaluate the prediction accuracy on each dataset. An AUC of $50\%$ represents the score achievable by random guessing. A high AUC score accounts for a high prediction performance. Results of the test AUC on all datasets are shown in Table~\ref{table:result}.

We compare the DKVMN model with the MANN baseline, the state-of-the-art DKT, the standard BKT model, and, when possible, optimal variations of BKT (BKT+). 
An interesting observation is that our implemented LSTM achieves better AUC than those in the original papers~\cite{piech2015deep,khajahdeep,xionggoing}. The reason may be that our implementations use norm clipping and early stopping, both of which improve the overfitting problem of LSTM. The results of BKT are directly obtained from recent works~\cite{khajahdeep,xionggoing}.

On the Synthetic-5 dataset, the DKVMN model achieves the average test AUC of 82.7\%. 
In our simulation, each exercise is treated as having a distinct skill label. MANN produces an average AUC of 81.0\%. DKT produces an AUC value of 80.3\%, which is better than the 75\% reported in the original paper~\cite{piech2015deep,khajahdeep}. BKT and its variant model achieve the AUC of 62\% and 80\% respectively~\cite{khajahdeep}. 
The prediction results of DKVMN from the ASSISTments2009 achieve improvement over MANN, DKT, and BKT with 81.6\% over 79.7\%, 80.5\%, and 63\% respectively~\cite{xionggoing}. As this dataset is preprocessed differently from that in~\cite{xionggoing}, their results are not comparable.
On the ASSISTments2015 dataset, the test AUC of DKVMN is 72.7\%, which is better than 72.3\% for MANN, 72.5\% for DKT (originally 70\% in~\cite{xionggoing}), and 64\% for classic BKT~\cite{xionggoing}.
With regard to Statics2011, which has the maximum number of exercise tags and the minimum number of answers, classical BKT gains the AUC of 73\% and BKT cooperating with forgetting, skill discovery, and latent abilities obtains an AUC of 75\%~\cite{khajahdeep}. Our implemented DKT leads to an AUC of 80.2\%, which is better than the 76\% from~\cite{khajahdeep}. MANN only produces the average AUC of 77.6\%. However, our DKVMN model achieves an AUC of 82.8\%, outperforming all previous models.

In summary, DKVMN performs better than other methods across all the datasets, particularly on the Statics2011 dataset whose number of distinct exercises is large. This result demonstrates that our DKVMN can model student's knowledge well when the number of exercises is very large.

DKVMN can achieve better prediction accuracy over student exercise performance and also requires considerably fewer parameters than the DKT model because of its large external memory capacity.
Table~\ref{table:result_KVMN_DKT} compares the DKVMN model with the DKT model using LSTM by traversing different hyperparameters. 
The table reveals that DKVMN with low state dimensions can achieve better prediction accuracy than DKT with high state dimensions.
For instance, on the Statics2011 dataset, DKT reaches the maximum test AUC of 80.20\% when the dimension of states equals 200 using 1 million parameters. Meanwhile, DKVMN can achieve the test AUC of 82.84\% only with 50 state dimensions using 197 thousand parameters.

Moreover, the DKT model suffers severe overfitting, whereas our DKVMN model does not confront such a problem.
As indicated in Figure~\ref{fig:AUC}, no huge gap exists between the training AUC and the validation AUC of DKVMN, and the validation AUC of DKVMN increases smoothly. However, as the epoch proceeds, the training AUC of DKT increases continuously, and the validation AUC of DKT only increases in the first several epochs and begins to decrease.

\begin{figure*}[tb!]
	\centering
	\includegraphics[width=0.9\textwidth]{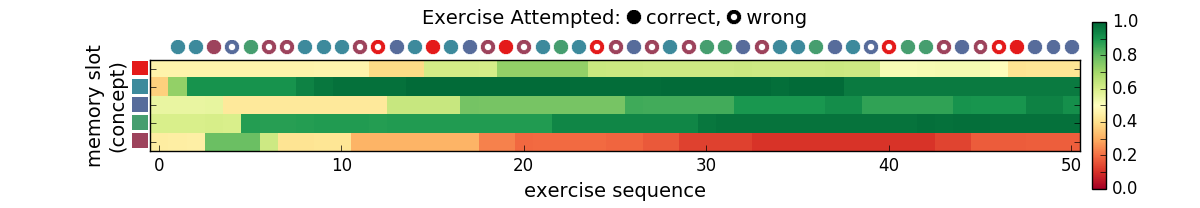}
	\caption{An example of a student's changing knowledge state on 5 concepts. Concepts are marked in different colors on the left side. After answering 50 exercises, the student masters the second, third, and fourth concepts but fails to understand the fifth concept. (Best viewed in color.)}
	\label{fig:memory_synthetic}
\end{figure*}

\subsection{Concept Discovery}
Our DKVMN model has the power to discover underlying patterns or concepts for exercises using the correlation weight $\textbf{w}$, which is traditionally annotated by experts.
The correlation weight between the exercise and the concept implies the strength of their inner relationship.
Compared with the conditional influence approach in~\cite{piech2015deep} which computes the dependencies between exercises and then defines a threshold to cluster the exercises, our model directly assigns exercises to concepts. No predefined threshold is required. As a result, our model can discover the concepts of exercises in an end-to-end manner.

Each exercise is usually associated with a single concept. In this case, we assign the exercise to the concept with the largest correlation weight value. From the experiments, we find that our model can intelligently learn sparse weight among concepts, and the discovered concepts reveal a compelling result.

On the Synthetic-5 dataset, each exercise is drawn from a concept $c^k$, where $k \in 1 ... 5$, such that the ground truth concept can be accessed for all exercises, as shown on the top x-axis of the heat map in Figure~\ref{fig:exe_cluster_synthetic}. Exercises from the same concept are labeled with squares in the same color.
The left heat map in Figure~\ref{fig:exe_cluster_synthetic} shows the correlation weight between 50 distinct exercises and 5 latent concepts (generated from DKVMN when the memory size is five). Each column represents the correlation weight between an exercise and five latent concepts. For each exercise, the weight is sparse where exactly one value approximates 1 and the others approximate 0. 
After clustering each exercise to the concept with the maximum weight value, we get the graph shown in the right part of Figure~\ref{fig:exe_cluster_synthetic}, which reveals a perfect clustering of five latent concepts. The adjusted mutual information~\cite{viola1997alignment} of our clustering result and the ground truth is 1.0. 

Moreover, when the memory size $N$ is set to be larger than the ground truth 5, e.g., 50, our model can also end up with 5 exercise clusters and find the appropriate concept for each exercise. Additional results are described in the appendix.

On the ASSISTments2009 dataset, no ground truth concept is used for each exercise. However, the name for each exercise tag can be obtained, as shown in the right part of Figure~\ref{fig:exe_cluster_assistments2009}. Each exercise tag is followed by a name.
The resulting cluster graph in Figure~\ref{fig:exe_cluster_assistments2009} is drawn using t-SNE~\cite{maaten2008visualizing} by projecting the multi-dimensional correlation weights to the 2-D points. 
All exercises are grouped into 10 clusters, where the exercises from the same cluster (concept) are labeled in the same color. The clustering graph reveals many reasonable results. Some related exercises are close to one another in the cluster.
For example, in the first cluster, $30$ {\em Ordering Fractions}, $33$ {\em Ordering Integers}, $37$ {\em Ordering Positive Decimals}, and $62$ {\em Ordering Real Numbers} are clustered together, which exposes the concept of {\em elementary arithmetic}. 
%

\subsection{Knowledge State Depiction}

Our DKVMN can also be used to depict the changing knowledge state of students. Depicting the knowledge state, especially each concept state, is helpful for the users on online learning platforms. If students possess their concept states of all concepts, which pinpoint their strengths and weaknesses, they will be more motivated to fill in the learning gaps independently. 
A student's changing knowledge state can be obtained in the read process using the following steps.

First, the content in the {\em value} component is directly used as the read content $\mathbf{r}_t$ in Eq.\eqref{eq:r_t}, which can be accessed by setting the correlation weight $\mathbf{w}_t$ to be $[0,..,w_i,..0]$, where $w_i$ of concept $c^i$ is equal to 1.

Then, we mask the weight of the input content embedding in Eq.\eqref{eq:f_t} to ignore the information of exercises:
\begin{equation}
\mathbf{f}_t = \text{Tanh}([\mathbf{W}_1^r, \textbf{0}]^T [\mathbf{r}_t,\mathbf{m}_{t}]+\mathbf{b}_1),
\end{equation}
where $W_1$ is split into two parts $\mathbf{W}_1^r$ and $\mathbf{W}_1^m$, and let $\mathbf{W}_1^m = \textbf{0}$.

Finally, we compute the scalar $p$ as in Eq.\eqref{eq:p_t} to be the predictive mastery level of a concept (concept state).

Figure~\ref{fig:memory_synthetic} shows an example of depicting a student's five changing concept states.
The first column represents the initial state of each concept before the student answers any exercise, such state differs from concept to concept.
Owing to our model's ability to discover concepts for each exercise, each time the student answers an exercise, the concept state of the discovered concept will increase or decrease. For example, when the student answers the first three exercises correctly, concept states of the second and fifth concepts increase; when the student answers the fourth exercise incorrectly, the concept state of the third concept decreases.
After answering 50 exercises, the student is shown to have mastered the second, third, and fourth concepts but failed to understand the fifth concept.

\section{Conclusions and Future Work}

This work proposes a new sequence learning model called DKVMN to tackle the KT problem. The model can be implemented in online learning platforms to improve the study efficiency of students. 
DKVMN not only outperforms the state-of-the-art DKT but can also trace a student's understanding of each concept over time, which is the main drawback of DKT. 
Compared with standard MANNs, the key-value pair allows DKVMN to discover underlying concepts for each input exercise and trace a student's knowledge state of all concepts.

For future work, we will incorporate content information into the exercise and concept embeddings to further improve the representations. We will also investigate a hierarchical key-value memory networks structure which can encode the hierarchical relationship between concepts.

\section{Acknowledgement}

The work described in this paper was partially supported by the Research Grants Council of the Hong Kong Special Administrative Region, China (No. CUHK 14208815 of the General Research Fund) and Ministry of Education of China (D.01.16.00101).

\section{Appendix}

\subsection{Concept Discovery}

When memory size $N$ is set to 50 for the synthetic-5 dataset, the exercises can still be clustered into five categories.
The heat map in Figure~\ref{fig:exe_cluster_synthetic50} describes all correlation weight vectors, which only fall into several concepts.
After clustering each exercise to the concept with the maximum weight value, the adjusted mutual information of our clustering result and the ground truth is 0.879. Additionally, if we cluster using t-SNE (in Figure~\ref{fig:exe_cluster_synthetic50}), then the adjusted mutual information will be 1.0, which reveals a perfect result.

\begin{figure}[!b]
	\centering
	\includegraphics[width=0.48\textwidth]{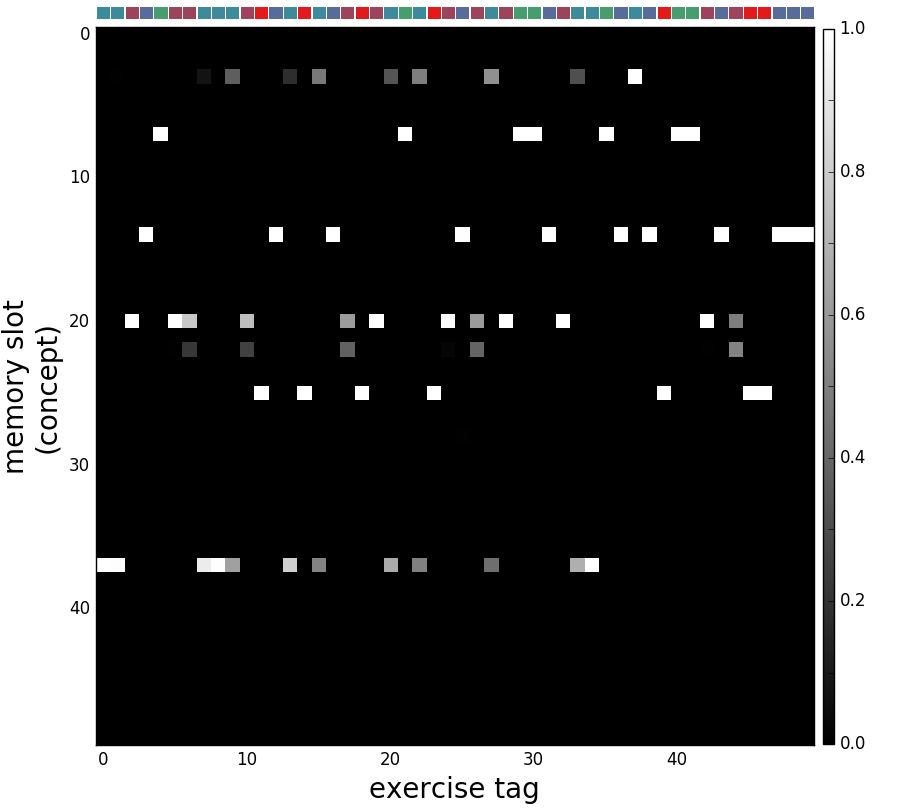}
	\includegraphics[width=0.09\textwidth]{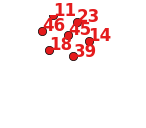}
	\includegraphics[width=0.09\textwidth]{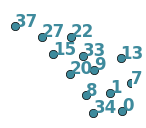}
	\includegraphics[width=0.09\textwidth]{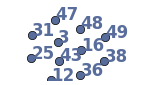}
	\includegraphics[width=0.09\textwidth]{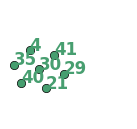}
	\includegraphics[width=0.08\textwidth]{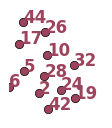}
	\caption{Concept discovery results on the synthetic-5 dataset when the memory size $N$ is set to 50. In the heat map, the x-axis represents each exercise and the y-axis represents the correlation weight between the exercise and five latent concepts generated from our DKVMN model. In the below exercise clustering graph using t-SNE, each node number represents an exercise. Exercises from the same ground-truth concept are clustered together. (Best viewed in color.)}
	\label{fig:exe_cluster_synthetic50}
\end{figure}

\subsection{Read Attention Mechanism of MANN}
For each input key $\mathbf{k}_t$, we compute the cosine similarity of the key and memory:
\begin{equation}
K[\mathbf{k}_t, \mathbf{M}_t(i)] = \frac{\mathbf{k}_t \cdot \mathbf{M}_t(i)}{\|\mathbf{k}_t\| \cdot \|\mathbf{M}_t(i)\|},
\end{equation}
which is then used to compute the read weight $\mathbf{w}^r$ through a softmax with a positive key strength $\beta_t$:
\begin{equation}
w_t^r(i) = \frac{\mathrm{exp}(\beta_t K[\mathbf{k}_t, \mathbf{M}_t(i)] )}{\sum_j \mathrm{exp}(\beta_t K[\mathbf{k}_t, \mathbf{M}_t(j)] )}.
\end{equation}

\subsection{Write Attention Mechanism of MANN}

The LRUA model~\cite{santoro2016meta} writes the keys either to the least used memory location or the most recently used memory location. 

First, a usage weight vector $\mathbf{w}_t^u$ is used to record the usage frequency of all memories:
The usage weights are updated at each time-step by decaying the previous usage weights and adding the current reading and writing weights:
\begin{equation}
\mathbf{w}_t^u = \gamma \mathbf{w}_{t-1}^u + \mathbf{w}_t^r + \mathbf{w}_t^w,
\end{equation}
where $\gamma$ is a decay parameter. $\gamma$ is fixed to be 0.9 in our implementation.
Then the least-used weight $\mathbf{w}_t^{lu}$ is defined to record the least-used memories using a notation $m(\mathbf{v},n)$, which denotes $n^{th}$ smallest element of the vector $\mathbf{v}$,
\begin{equation}
w_t^{lu}(i) = 
\begin{cases}
0       & \quad \text{if } w_t^u(i) > m(\mathbf{w}_t^u,n)\\
1       & \quad \text{if } w_t^u(i) \leq m(\mathbf{w}_t^u,n),
\end{cases}
\end{equation}
where $n$ is set to equal the number of reads to memory.

Now the write weight $\mathbf{w}_t^w$ is the convex combination of the previous read weights and previous least-used weights:
\begin{equation}
\mathbf{w}_t^w = \sigma(\alpha) \mathbf{w}_{t-1}^s + (1-\sigma(\alpha) ) \mathbf{w}_{t-1}^{lu},
\end{equation}
where $\sigma(\cdot)$ is a sigmoid function, and $\alpha$ is a scalar gate parameter to interpolate between two weights.

\newpage

%
\bibliographystyle{abbrv}
\bibliography{sigproc}  

\end{document}